# Formalizing Scenario Analysis

**Peter McBurney**
Department of Computer Science
University of Liverpool
Liverpool L69 7ZF UK
p.j.mcburney@csc.liv.ac.uk

**Simon Parsons**
Sloan School of Management
Massachusetts Institute of Technology
Cambridge MA USA
sparsons@mit.edu

## Abstract

We propose a formal treatment of scenarios in the context of a dialectical argumentation formalism for qualitative reasoning about uncertain propositions. Our formalism extends prior work in which arguments for and against uncertain propositions were presented and compared in interaction spaces called Agoras. We now define the notion of a scenario in this framework and use it to define a set of qualitative uncertainty labels for propositions *across* a collection of scenarios. This work is intended to lead to a formal theory of scenarios and scenario analysis.

## 1 Introduction

In many domains, the absence of hard data or the presence of conflicting perceived interests makes reaching agreement on the quantification of uncertainty difficult. Argumentation formalisms have been proposed for the qualitative representation of uncertainty in these circumstances (Krause *et al.* 1995) and have found application in intelligent systems, for example in medical and safety analysis domains (Carbogim, Robertson, & Lee 2000). In (McBurney & Parsons 2000), we proposed a formalism using dialectical argumentation for representing and resolving the arguments for and against uncertain propositions. This representation was grounded in specific theories of rational human discourse and was centered on an electronic space for presentation of arguments, which we termed an *Agora*. In subsequent work (McBurney & Parsons 2001b), we extended this formalism and demonstrated that it had several desirable properties when used for inference and decision-making. In this paper, we further extend this framework to enable dialectical argumentation under and between multiple circumstances, or scenarios.

The notion of scenario (Schwartz 1991) has found widespread application in business forecasting, in public policy determination, and in scientific domains. An early use of the methods of scenario analysis may be seen in nineteenth-century statistical mechanics, where research sought to determine if the properties of a physical system, such as its entropy at a given time, depended on the system's initial state. Ludwig Boltzmann (Boltzmann 1872) tackled this problem by comparing the given system to a collection of alternative, imaginary systems, each having different initial conditions — i.e., what we would now call scenarios. By doing so, he could potentially assess the extent to which the system property of interest was independent of the initial system state. Josiah W. Gibbs (Gibbs 1902) formalized the concept of a collection of alternative systems with his notion of *ensemble*, a term we also use.

Perhaps the most important and complex recent application of scenario analysis has been in the work of the Intergovernmental Panel on Climate Change (IPCC) (McCarthy *et al.* 2001), the UN agency tasked with assessing the current and possible future states of the world's ecosystem, and with considering and recommending appropriate environmental regulatory policies. In this domain, scenario analysis has been used for scientific modeling and prediction, for the modeling of socio-economic variables and conditions, and for the assessment of proposed regulatory policies and targets (Carter & others 2001).

Despite their widespread use, however, there appears to be no formal theory of scenarios or scenario analysis. Without a formal theory, many questions remain without rigorous answers, e.g., How should scenarios be constituted? How many scenarios should be considered? How should individual scenarios be analysed? How should any differences in the likelihood of occurrence of different scenarios be represented? How should their relative importance be represented? How should reasoning be undertaken *across* a collection of scenarios, or multiple collections of scenarios? In the absence of a formal theory of scenarios it is difficult to assess the validity or reliability of any particular application of scenario analysis, for example, the many analyses generated by the work of the IPCC (Carter & others 2001). Moreover, because no computational theory of scenarios yet exists, application of scenario analysis in intelligent systems is limited.

The long-term aim of the research reported here is a rigorous, formal, computational theory of scenarios. In this paper, we take an initial step towards this aim, by considering one type of scenario, those based on dialectical argumentation systems. In Section 2, we review our model for qualitative inference in uncertain domains, which uses dialectical argumentation to represent conflicting, ambiguous or contested information. Section 3 defines our notions of scenario and ensemble, while Section 4 considers the question of when two scenarios may be considered distinct. Section 5 then considers how many observations we need to take for any debate in order to estimate its long term position; here we prove what we believe to be an important theorem, Proposition 4, which says that the most recent snapshot of a debate is at least as good, in a precise sense, as any combination of earlier snapshots. In Section 6, we consider the the assignment of uncertainty labels to uncertain propositions on the basis of their argumentation status in a collection of scenarios. This is illustrated with an example in Section 7 and the paper concludes with a discussion of related work in Section 8.

One possible response to these proposals is that scenario analysis is unnecessary in an argumentation context, because these frameworks have been developed precisely to represent conflicting or uncertain information, and to resolve any inconsistency in the resulting conclusions. Such a view is mistaken. In a typical application, we are not merely trying to decide whether the possible arguments for some given claim are, on balance, stronger or weaker than the arguments against it; we are also trying to identify the *circumstances* (the assumptions and allowed rules of inference) under which arguments exist for or against the claim, and the *circumstances* under which those arguments for it are stronger than those against it. To do this rigorously, we need to clearly demarcate the sets of possible circumstances — i.e., the scenarios — from one another and to compare them.

## 2 Dialectical Argumentation

In this section we briefly summarize the Agora framework for the qualitative representation of uncertainty presented in (McBurney & Parsons 2000; 2001b). In this framework, arguments for and against claims are articulated by participants in an electronic space, called an *Agora*, with claims expressed as formulae in a propositional language. By means of defined locutions, participants in the Agora can variously posit, assert, contest, justify, rebut, undercut, qualify and retract claims, just as happens in real discourse. For example, a debate participant $\mathcal{P}_i$ could demonstrate her argument $\mathcal{A}(\rightarrow \theta)$ supporting a claim $\theta$, an argument to which she was committed with strength $D$, by means of the locution:

$$\text{show\_arg}(\mathcal{P}_i : \mathcal{A}(\rightarrow \theta, D)).$$

The rules governing the use of each permitted locution are expressed in terms of a formal dialogue-game between the participants (Hamblin 1970). We assume that the Agora participants begin a debate with a set of agreed facts, or assumptions, and an agreed set of inference rules. Because we want to model many forms of reasoning, these rules need not be deductive and may themselves, in our Agora formulation, be the subject of argument.

We demonstrated the use of this framework for the representation of uncertainty by defining a set of uncertainty labels assigned to claims on the basis of the arguments presented for and against them in the Agora. Essentially, one could say that claims have more credibility (and hence less uncertainty) the fewer and the weaker are the arguments against them. While any set of labels could be so defined, we drew on earlier work in argumentation (Krause *et al.* 1995) and defined the set: {*Accepted, Probable, Plausible, Supported, Open*}, with the elements listed in decreasing order of certainty. For example, a claim was regarded as *Probable* at a particular time if at least one consistent argument had been presented for it in the Agora by that time, but no arguments for its negation (rebuttals) nor for the negation of any of its assumptions (undercuts) had been presented by then. We defined a claim as *well-defended* at a given time if there was an argument for it at that time and any rebuttals or undercuts had themselves been subject to counter-rebuttals or to undercuts. *Accepted* claims at any time were defined as those which are well-defended at that time.[1]

As arguments for and against a proposition are presented to the Agora, the status of a proposition may rise and fall: a claim considered *Probable* at one time may be only *Plausible* later, and then be *Accepted* later again. We therefore defined the truth-valuation of a claim $\theta$ at time $t$, denoted $v_t(\theta)$, to be 1 if $\theta$ had the label *Accepted* at this time, otherwise it was 0. Such a valuation summarizes the knowledge of the community of debate participants at the particular time, since it incorporates, via the definitions of the labels, all the arguments for and against $\theta$ articulated to that time. Consequently, assessing the truth-status of a claim at a particular time can be viewed as taking a *snapshot* of an Agora debate. Of course, because these definitions are time-dependent, and arguments may be articulated in the Agora at any time, such an assignment of uncertainty labels and truth valuation must be defeasible. Claims accepted at one time may be overturned at another, in the light of new information learnt or arguments presented subsequently.

In using the Agora framework to represent uncertainty, at-tention will focus on the truth valuation function over the long-run.[2] The sequence $(v_t(\theta) \mid t = 1, 2, \ldots)$ may or may not converge as $t \to \infty$. Suppose that it does converge, and denote its limit value by $v_\infty(\theta)$. What will the value of a snapshot taken at time $t$, namely $v_t(\theta)$, tell us about $v_\infty(\theta)$? Of course, since any finite snapshot risks being overtaken by subsequent information or arguments, we cannot infer with complete accuracy from the finite snapshot to the infinite value. However, we have shown (McBurney & Parsons 2001b) that, under certain conditions, we can place a bound on the likelihood that such an inference is in error. The conditions essentially require that: (a) the snapshot is taken at a time after commencement sufficient for all the arguments using the initial information to have been presented, and (b) there is a bound on the probability that new information arises following the snapshot. This result is proved as Proposition 7 of (McBurney & Parsons 2001b), which we reproduce here. For this, we first need some definitions.

---

[1] These labels are assigned on the basis of the arguments presented by all participants in the Agora; thus, individual participants may not agree with any label assignment, since their own arguments will typically only be a subset of those presented.

**Definition 1:** *We write $LE_\theta$ for the statement:* "The function $v_t(\theta)$ converges to a finite limit as $t \to \infty$." *We also write $\mathcal{X}_{t,\theta}$ for the statement:* "New information relevant to $\theta$ becomes known to an Agora participant after time $t$."

In general, at any time $s$, we do not know whether new evidence will become available to Agora participants at a later time $t$ or not. Consequently, the variables $\mathcal{X}_{t,\theta}$, for $t$ not in the past, represent uncertain events. Also uncertain for the same reason are statements concerning the future values of $v_t(\theta)$ for any $\theta$. Because these events are uncertain, we assume the existence of a probability function over them, i.e., a real-valued measure function mapping such statements to $[0, 1]$ which satisfies the axioms of probability.

**Definition 2:** *$Pr(.)$ is a probability function defined over statements of the form $\mathcal{X}_{t,\theta}$ and statements concerning the values of $v_t(\theta)$, for any formula $\theta$.*

**Proposition 1:** [Proposition 7 of (McBurney & Parsons 2001b)] *Let $\theta$ be a formula and suppose that all arguments pertaining to $\theta$ and using the information available at commencement are articulated by participants by some time $s > 0$. Suppose further that $v_{t_m}(\theta) = 1$ for some $t_m \geq s$. Also, assume that $Pr(\mathcal{X}_{t_m,\theta}) \leq \epsilon$, for some $\epsilon \in [0, 1]$. Then the following inequalities hold:*

$$Pr(LE_\theta \text{ and } v_\infty(\theta) = 1 \mid v_{t_m}(\theta) = 1) \geq 1 - \epsilon.$$

$$Pr(LE_\theta \text{ and } v_\infty(\theta) = 0 \mid v_{t_m}(\theta) = 1) \leq \epsilon. \qquad \square$$

Like the standard (Neyman-Pearson) procedures for statistical hypothesis testing, this proposition provides us with some confidence in our use of finite snapshots to make inferences about the long-run truth-valuation function for a debate. While such inference is not deductively valid, at least its likelihood of error may be bounded.[3] In the sections below, we will be comparing the results of debates in more than one Agora. We therefore assume that we have a single probability function $Pr$ defined across all the relevant statements. We will also index symbols with superscripts ([1],[2], etc) to denote the Agora to which they refer. We next define the concept of Scenario.

## 3 Scenarios and Ensembles

The framework we have just outlined provides a means to represent the diverse arguments that may be derived from a given set of assumptions, by means of a given set of inference rules (deductive or otherwise). If we were to start with a different set of assumptions, and/or permit the use of different inference rules, the arguments presented in the Agora could well be different. As a result, the uncertainty labels and truth values assigned to formulae could also be different, both when taken at finite snapshots and in the limit. Each collection of alternative sets of assumptions and inference rules we call a scenario, which we define as follows:

**Definition 3:** *A **Scenario** for a given domain consists of a set of assumptions and a set of inference rules, with which participants are equipped at the commencement of an Agora debate over propositions in that domain. We denote scenarios for a given domain by $S^1, S^2, \ldots$, etc. For each scenario, $S^i$, an Agora debate undertaken with the assumptions and inference rules of that scenario, is said to be the **associated Agora**, denoted $\mathcal{A}^i$. We assume only one debate is conducted in association with any scenario.*

Because we wish to reason across multiple scenarios, we also define:

**Definition 4:** *An **Ensemble** $\mathcal{S}$ is a finite collection of distinct Scenarios $\{S^1, \ldots, S^m\}$ relating to a common domain. We assume that, associated with each scenario $S^i \in \mathcal{S}$, is a real-number, $a^i \in [0, 1]$, called its **scenario weight**. We call $\bar{a} = (a^1, a^2, \ldots, a^m)$ the **ensemble weights vector** of $\mathcal{S}$.*

We do not assume the weights sum to unity across the $m$ scenarios, although they may do so. The weights may vary with time, but, if so, we assume that their assignment to scenarios is independent of the dialectical status of claims in the corresponding debates. This assumption is made because the assignment of weights to scenarios should be on the basis of characteristics of the scenarios themselves, not on the basis of arguments which ensue or don't ensue in the associated Agora debates.

---

[2] Strictly, we are assuming throughout that time in the Agora is discrete, and can be represented by a countably-infinite set.

[3] One may object that we can never know the value of $\epsilon$. While this is true, participants in a debate are often quite willing to provide subjective estimates for such probability bounds. Scientists, for example, will often estimate the chance that new information will emerge in future which will overturn an established theory.

What interpretation we give to the weights depends upon the meanings we give to the logical language, to the scenarios and to arguments for claims in the corresponding Agora debates. For example, the assumptions and claims may represent objects in the physical world, and the inference rules physical manipulations of these objects, such as actual construction of new objects from existing ones. Scenarios can thus be interpreted as different sets of resourcing assumptions, with claims being well-defended in an Agora debate when the objects they represent are able to be constructed with the assumed resources. In this interpretation, the weights attached to scenarios may be the relative costs or benefits of different resources, or their likelihoods of occurrence. A second interpretation could arise where the scenarios represent alternative sets of rules of procedure for interaction between a group of participants, for example, in a legal domain or in automated negotiation. Here the rules of inference may represent different allowable modes of reasoning, such as reasoning by analogy or from authority. The weights may represent the extent of compliance of each scenario with some set of principles of rational discourse, such as those of (Hitchcock 1991), or with some normative economic or political theory. Finally, a third interpretation would have the scenarios as different descriptions of some uncertain domain, for example different scientific theories, with propositions being statements about the domain, and the inference rules representing different causal mechanisms. The scenario weights could be relative likelihoods of occurrence, or valuations of relative importance or utility. This third interpretation is the one we will consider in this paper.

## 4 Comparing Scenarios

### 4.1 Comparing two long-run debates

Our definition of an Ensemble says that the scenarios included must be distinct. We require this so that when aggregating across scenarios we do not engage in "double-counting" of separate scenarios which are really the same. When are two scenarios the same? Obviously, we may consider them to be the same when they have identical sets of assumptions and inference rules. But two scenarios identical in this fashion may result in very different Agora debates, as different arguments may be presented in each, or the same arguments may be presented at different times. It is not clear, therefore, that identical scenarios will lead to identical assignments of truth-labels, even over the long-run; we show that, under certain conditions, they will do so. Throughout this section $S^1$ and $S^2$ will be two scenarios of interest, and $\mathcal{A}^1$ and $\mathcal{A}^2$ their associated Agora debates.

**Proposition 2:** *Let $\theta$ be a claim. Suppose that $S^1$ and $S^2$ are identical scenarios, i.e., they have identical sets of assumptions and identical sets of inference rules. Suppose that in the corresponding Agora debates, $\mathcal{A}^1$ and $\mathcal{A}^2$, all possible arguments based on the initial assumptions and using the inference rules are eventually articulated. Suppose further that no new information is presented to either debate following commencement. Then, the long-run truth status of $\theta$ in each debate is the same.*

**Outline of Proof:** Given the premises, the only way the two debates will potentially differ will be in the order that arguments are articulated. But if all arguments are eventually articulated, then after some finite time no further arguments will be presented in either debate. The definitions of the truth valuation functions in (McBurney & Parsons 2001b) depend only the arguments which have been presented at any time, and not their order of presentation. The proposition follows.  □

If we relax the assumption that no new information arrives in either debate our conclusion acquires a probabilistic qualification. While this does not guarantee that two identical scenarios always lead to identical long-run truth assignments, it does bound the likelihood that such is not the case.

**Proposition 3:** *Let $\theta$, $S^1$ and $S^2$ be as before. Suppose there exist upper bounds $\epsilon^i \in [0, 1]$ for the probability that new information arrives after commencement in debate $i$, i.e., that $Pr(\mathcal{X}_{0,\theta}^i) \leq \epsilon^i$, for $i = 1, 2$. Then we have:*

$$Pr(v_\infty^1(\theta) = v_\infty^2(\theta)) \geq 1 - \epsilon^1 - \epsilon^2.$$

**Outline of Proof:** By the previous result, the two long-run assignments of truth to $\theta$ are only different if one or other debate receives new information. The probability that this occurs is less than or equal to the sum of the probabilities that either debate receives new information less the probability that they both do. This latter event has probability greater than or equal to zero, and the inequality follows by algebraic manipulation.  □

### 4.2 A decision rule for scenario comparison

We now provide a decision rule for determining if two scenarios $S^1$ and $S^2$ are the same. This decision rule classifies scenarios into two classes, labeled *distinct* and *non-distinct*. The rule proposed for determination of distinctness of scenarios uses two criteria (in order of application): (a) whether or not the two scenarios have identical assumptions and inference rules; (b) in the case where they do, whether or not either scenario is judged to have a high probability of receiving new information.

**Case 1:** $S^1 \neq S^2$. Conclude that the two scenarios are distinct.

**Case 2A:** $S^1 = S^2$ and $Pr(\mathcal{X}_{0,\theta}^1), Pr(\mathcal{X}_{0,\theta}^1)$ **both small.** In this case, the likelihood of new information arising in either scenario is small and Proposition 3 allows us to infer that $v_\infty^1(\theta) = v_\infty^2(\theta)$ with high probability. Conclude that the two scenarios are *non-distinct*.

**Case 2B:** $\mathcal{S}^1 = \mathcal{S}^2$ **and one or both of** $Pr(\mathcal{X}_{0,\theta}^1), Pr(\mathcal{X}_{0,\theta}^1)$ **large.** In this case, the likelihood of new information arising in at least one scenario is large, and thus, Proposition 3, it is unlikely that $v_\infty^1(\theta) = v_\infty^2(\theta)$. Conclude that the two scenarios are *distinct*.[4]

In the first case, where the two scenarios have different premises and/or inference rules, we classify them as distinct. Two such distinct scenarios, of course, may result in the same arguments being presented in both scenarios after some finite time. In the other two cases (Cases 2A and 2B), where the underlying assumptions and inference rules are the same in the two scenarios, Proposition 2 says that the long run truth assignments for $\theta$ in the corresponding Agora debates, if they exist, will be identical, provided no new information is presented in either Agora debate following commencement. If new information is presented, then Proposition 3 provides a bound for the probability that the long-run truth assignments are the same, in terms of the probabilities of new information being received. In the case (Case 2A) when these probabilities are believed to be small, the two long-run truth assignments are most likely identical, and we can classify the two scenarios as being the same. In the other case (Case 2B), where one or both probabilities are large, we classify the two scenarios as not the same.

Note that, although under Cases 2A and 2B we are making inferences about the long run truth assignments, $v_\infty^1(\theta)$ and $v_\infty^2(\theta)$, these inferences are based only on the premises and inference rules used and assessments of the probability of new information being received after commencement of the associated Agora debates. These inferences, and hence this classification, do not depend on the progress or status of the debates themselves. In other words, our classification of scenarios is not based on the output of the debates conducted under the scenarios.

## 5　Observing Agora debates

What may we feasibly observe about an Agora debate? Firstly, we could take a snapshot at a particular finite time after commencement. Or, secondly, we could take a number of such snapshots. Or, thirdly, we could examine the actual arguments used in a debate from commencement up to a particular time. In the first subsection below, we show that taking the most recent snapshot is at least as good an indicator of the long run status of a debate as any other combination of earlier snapshots. In other words, we need only take one snapshot to capture all the information available in a debate. In the second subsection we consider how we may compare a snapshot from one debate with that from another. The third approach — considering the arguments themselves — we leave for another occasion. As before, we denote the long-run truth status of a formula $\theta$ in debate $i$, if this limit exists, by $v_\infty^i(\theta) = \lim_{t\to\infty} v_t^i(\theta)$. The subsections which follow will discuss finite estimators of this long-run value, estimators we denote by $\hat{v}_\infty^i(\theta)$.

### 5.1　Observing a debate

Suppose that we have multiple snapshots of a debate, i.e., for a debate $\mathcal{A}$ we have a sequence of observations of the truth-status of $\theta$: $v_1(\theta), v_2(\theta), \ldots, v_n(\theta)$. These values are all either zero or one, and each is an estimate for the long-run truth status $v_\infty(\theta)$ of $\theta$. Given such multiple estimates, there are a number of ways we could combine them to produce a single estimate of $v_\infty(\theta)$, for example: (a) **The mean**, which is the sum of the observed snapshot values, divided by $n$; (b) **An $\alpha, \beta$-trimmed mean**, which is the mean calculated after first ranking the observations in ascending order and then deleting $\alpha$ % of the observations at the lower end and $\beta$ % at the upper end (Huber 1981). For instance, we may delete those at the beginning of the sequence, on an assumption that early values of $v_t(\theta)$ will oscillate as all the relevant arguments are presented to the Agora; (c) **the mode**, the most common value, i.e., whichever of 0 or 1 appears most frequently in the $n$ observations; etc.

However, each truth-valuation $v_k$ is defined in terms of the arguments presented to the Agora up to time $k$, so, in some sense, each observation summarizes all the information relevant to $\theta$ up to and including the time the observation was made. We should therefore expect the final observation, $v_n$ to contain the most information, and so to be the best estimator (in some sense) of the long-run value, $v_\infty$.[5] This is indeed the case, as the following theorem shows. For simplicity, we omit $\theta$ from the notation.

***Proposition 4:*** *Let $v_1, v_2, \ldots, v_n$ be a sequence of $n$ snapshot values concerning $\theta$ taken from a debate $\mathcal{A}$. Suppose the limit $v_\infty = \lim_{n\to\infty} v_n$ exists. Let $\hat{v}_n$ be the estimator of $v_\infty$ using only the final observation in such a sequence. Further, let $\hat{v}_n^*$ be any estimator of $v_\infty$ based on these $n$ observations which converges to a finite limit as $n \to \infty$. Then:*

$$\lim_{n\to\infty} Pr(|\hat{v}_n - v_\infty| \leq |\hat{v}_n^* - v_\infty|) = 1.$$

**Proof.** We prove this by contradiction. If the result does not hold, then there must exist $\epsilon > 0$ such that:

$$\lim_{n\to\infty} Pr(|\hat{v}_n - v_\infty| > |\hat{v}_n^* - v_\infty|) = \epsilon$$

---

[4] Note that the meanings of *"small"* and *"large"* may be domain dependent.

[5] Note that the final observation is itself an $\alpha, \beta$-trimmed mean.

Then, there must be infinitely many $m$ such that $|\hat{v}_m - v_\infty| > |\hat{v}_m^* - v_\infty|$. Since the sequence of final values $v_n$ is a sequence of zeros or ones, and it converges to $v_\infty$, then for each of these $m$, we have one of two cases:

[$v_\infty = 1$:] By the strict inequality, we must have $\hat{v}_m = 0$.

[$v_\infty = 0$:] Likewise, we must have $\hat{v}_m = 1$

But this happens for infinitely many values $m$, which contradicts the assumption that $v_n$ converges to $v_\infty$. □

This result shows that the final observation of a sequence of snapshots is at least as good, in the long run, as any other convergent estimator of $v_\infty$ based on this sequence. We therefore need only consider the most recent snapshot in any assessment of the truth status of a claim in a debate.

### 5.2 Comparing two debates

We now consider the comparison of two Agora debates, undertaken under different scenarios, by means of finite snapshots of each at a particular time $t$ after commencement. Each snapshot, $v_t^i(\theta)$, will give us an assessment of the long-term truth-status of a claim $\theta$ in each debate $\mathcal{A}^i$, for $i = 1, 2$. That is, we set $\hat{v}_\infty^i(\theta) = v_t^i(\theta)$. Proposition 1 tells us that we can bound the probability of error in inferring from the finite estimate $\hat{v}_\infty^i(\theta)$ to the true infinite limit value $v_\infty^i(\theta)$. Can we also bound the probability of error when inferring from a comparison of the finite values? The next proposition provides such bounds.

***Proposition 5:*** *Let $\theta$ be a formula, and let $\mathcal{A}^1, \mathcal{A}^2$ be two Agora debates associated with scenarios $\mathcal{S}^1, \mathcal{S}^2$ respectively. Suppose that, in debate $\mathcal{A}^i$ ($i = 1, 2$), all arguments pertaining to $\theta$ and using the information available at commencement are articulated by participants by some time $s_i > 0$. Suppose further that there is a time $m \geq max(s_1, s_2)$, such that $v_m^1(\theta) = v_m^2(\theta)$. Also, assume that there exist $\epsilon_1, \epsilon_2 \in [0, 1]$, possible dependent on $m$, such that each $Pr(\mathcal{X}_{m,\theta}^i) \leq \epsilon_i$. Then the following four inequalities hold:*

1. $Pr(LE_\theta^1$ and $LE_\theta^2$ and $v_\infty^1(\theta) = v_\infty^2(\theta) \mid v_m^1(\theta) = v_m^2(\theta)) \geq 1 - \epsilon_1 - \epsilon_2$.

2. $Pr(LE_\theta^1$ and $LE_\theta^2$ and $v_\infty^1(\theta) \neq v_\infty^2(\theta) \mid v_m^1(\theta) = v_m^2(\theta)) \leq \epsilon_1 + \epsilon_2$.

3. $Pr(LE_\theta^1$ and $LE_\theta^2$ and $v_\infty^1(\theta) = v_\infty^2(\theta) \mid v_m^1(\theta) \neq v_m^2(\theta)) \leq \epsilon_1 + \epsilon_2$.

4. $Pr(LE_\theta^1$ and $LE_\theta^2$ and $v_\infty^1(\theta) \neq v_\infty^2(\theta) \mid v_m^1(\theta) \neq v_m^2(\theta)) \geq 1 - \epsilon_1 - \epsilon_2$.

**Proof.** Arguments similar to Proposition 3. □

Using simultaneous finite snapshots of two debates to make an inference about the long-run truth-status of a formula is a process prone to error. This result says that, under certain circumstances, we can bound the probability of such errors. The "certain circumstances" relate to the timing of the snapshots — which must be long enough into the two debates for all the arguments based on the initial information to have been presented — and to the probabilities of new information being presented to each debate subsequent to the snapshots being taken. As one would expect, the error bounds are functions of these probabilities.[6]

In proving this result, we have not assumed that the event of new information being presented to one Agora debate is independent of new information being presented to the other. If we were able to make such an assumption, our error bounds would be tighter, with the product $\epsilon_1 \epsilon_2$ added to the right-hand side of the first and fourth inequalities and subtracted from the second and third. Independence of these two events is a function of how "distinct" are the two scenarios. Scenarios in the same domain which are very similar are likely to experience new information concurrently; participants in the corresponding debates are also likely to make similar assessments of the relevance of such new information.

## 6 Reasoning across Scenarios

Scenario analysis is typically used to answer one or more questions about an application domain. Users may wish to know whether some proposition $\theta$ is true under any scenario at all, e.g., *Is $\theta$ possible?* If it is possible, they may wish to then know the proportion of scenarios in which this is the case, e.g., *How likely is $\theta$?* Indeed, in the extreme all scenarios will be considered in order to answer the question, *Is $\theta$ inevitable?* Guided by these questions, we now define a set of qualitative labels to express the truth status of a claim $\theta$ across multiple, distinct scenarios. To do this, we assume throughout this section that we have an ensemble $\mathcal{S} = \{\mathcal{S}^1, \mathcal{S}^2, \ldots, \mathcal{S}^m\}$, each $\mathcal{S}^i$ with an associated Agora debate $\mathcal{A}^i$, and with associated scenario weight $a^i$. We assume $\theta$ is some uncertain proposition and we denote the truth valuation of $\theta$ in Agora debate $\mathcal{A}^i$ at time $t$ by $v_t^i(\theta)$, for $i = 1, \ldots, m$.

***Definition 5:*** *Given an ensemble $\mathcal{S}$ and a proposition $\theta$, we define the **Ensemble support** for $\theta$ at time $t$ by*

$$m_t^{\mathcal{S}}(\theta) = \frac{\sum_{i=1}^m a^i v_t^i(\theta)}{\sum_{i=1}^m a^i}.$$

---

[6]One could also view each sequence as arising from a random process, and so view the comparison problem as a test of an hypothesis that the two sequences are governed by the same probability distribution. Because the form of the distributions is not specified, the appropriate test would be nonparametric, e.g., the Kolmogorov-Smirnov two-sample test (Gibbons 1985). However, the asymptotic theory for even these tests requires that the underlying distributions be continuous and that the two samples be drawn independently. Neither assumption is appropriate here.

Given a fixed real number $\epsilon \in (0, 0.5)$, we now define various classes of support, as follows:

- $\theta$ is said to be **Inevitable** at time $t$ precisely when $m_t^S(\theta) = 1$. This class of propositions is denoted $\mathcal{A}_{I,t}$.

- $\theta$ is said to be $100(1-\epsilon)\%$-**Certain** at time $t$ precisely when $m_t^S(\theta) \geq 1 - \epsilon$. This class is denoted $\mathcal{A}_{1-\epsilon,t}$.

- $\theta$ is said to be **Probable** at time $t$ precisely when $m_t^S(\theta) \geq 0.5$. This class is denoted $\mathcal{A}_{P,t}$.

- $\theta$ is said to be $100\epsilon\%$-**Possible** at time $t$ precisely when $m_t^S(\theta) \geq \epsilon$. This class is denoted $\mathcal{A}_{\epsilon,t}$.

- $\theta$ is said to be **Open** at time $t$ if it is a well-formed formula of the logical language over which Agora debates are conducted. This class is denoted $\mathcal{A}_{O,t}$.

Following from these definitions, we have an inclusion hierarchy on these classes of propositions:

**Proposition 6:** *For a fixed $\epsilon \in (0, 0.5)$ and time $t$,*

$$\mathcal{A}_{I,t} \subseteq \mathcal{A}_{1-\epsilon,t} \subseteq \mathcal{A}_{P,t} \subseteq \mathcal{A}_{\epsilon,t} \subseteq \mathcal{A}_{O,t}. \qquad \square$$

We can then assign qualitative labels to any proposition $\theta$ according to which of these sets it belongs to. Note that in any one Agora debate, arguments may be presented for both $\theta$ and for $\neg\theta$, and, indeed, it is possible for both propositions to be well-defended in the same debate simultaneously.[7] Thus it is not necessarily the case, in this framework, that there is any complementarity between the uncertainty label assigned to a proposition $\theta$ and the label assigned to its negation; both propositions may be assigned *Probable*, for example, or both *75%-Certain*, depending on the arguments which support them.

## 7 Example

Given space limitations, our example is very simplified, and really only illustrates the concept of assigning uncertainty labels across an ensemble of scenarios.[8] We consider the situation facing an intending operator of global mobile satellite-based telecommunications services (GMSS) in 1990 (McBurney & Parsons 2002). Demand for these services was predicted to depend heavily on the extent to which terrestrial mobile communications services would expand, both in terms of customer numbers and the geographic area under coverage. One could imagine a number of scenarios for the future, under each of which there would be arguments for and against the claim that demand for GMSS would be large. We consider an ensemble of three scenarios, which can readily be seen to be distinct:

**Scenario 1:** Terrestrial mobile services expand rapidly and customers wish to use their phone everywhere, both inside and outside terrestrial coverage. Argument: Large numbers of terrestrial customers leads to high demand for GMSS outside terrestrial coverage.

**Scenario 2:** Terrestrial mobile services expand rapidly and customers are happy with the terrestrial coverage, not wishing to use it outside. Argument: Large geographic coverage for terrestrial services leads to low demand for GMSS, as most of the world has coverage.

**Scenario 3:** Terrestrial mobile services do not expand, but customers wish to use their phone everywhere, both inside and outside terrestrial coverage. Argument: Small geographic terrestrial coverage means high demand for GMSS.

Recall that scenario weights are assigned independently of the arguments under each scenario; assume this ensemble has a weights vector of the likelihoods $(0.7, 0.7, 0.3)$. Suppose $\epsilon = 0.05$. Let $\theta$ be the claim: *"GMSS experiences high demand."* There are well-defended arguments for $\theta$ in Scenarios 1 and 3, which have respective weights of 0.7 and 0.3. Thus we can say that $\theta$ is *5%-Possible*. Since $(0.7 + 0.3)/(0.7 + 0.7 + 0.3) = 0.59$, we can also conclude that $\theta$ is *probable*, but not that it is *95%-Certain* or *Inevitable*. We have thus assigned an uncertainty label to the proposition $\theta$, on the basis of its argumentation status within each scenario and taking into account the assumed relative weights of each scenario.

## 8 Discussion

Despite their widespread use, there is as yet no formal, computational theory of scenarios and scenario analysis. In this paper, we have commenced work on such a theory for scenarios which describe debates over uncertain propositions. In our formalism a scenario is a set of specified premises and inference rules, which participants to a debate use to engage in argument. We have presented a rule for determining whether two such scenarios are distinct or not, based only on their respective premises and inference rules, and on estimates of the probability that each debate will receive new information in the future. We have also shown that, when using finite snapshots of a debate to estimate the long-run truth status of a proposition, it is suf-ficient to use the most recent snapshot; this is at least as good (in a precise sense) as using earlier snapshots. We then defined a set of qualitative uncertainty labels for the truth status of claims when debates have been conducted under multiple, distinct scenarios. These labels provide a means to aggregate across the scenarios in a formal manner. Assigning weights to the scenarios — for example, to represent their relative probabilities of occurrence — enables the aggregate-level labels to be used to predict the truth-status of claims in the world beyond the debates.

---

[7]This is essentially because each claim may be defended by arguments which do not attack each other. An interesting question is under what circumstances the Ensemble Support function satisfies the axioms of probability. We conjecture that this is so if the attack relationship between arguments only permits rebuttals and not undercuts, thereby ensuring that every argument which attacks a claim is also an argument for its negation.

[8]In particular, we do not illustrate the working of the argumentation apparatus over time within the Agora debate under each scenario, since this is presented in (McBurney & Parsons 2001b).

The work presented here is novel. The closest work we have found used cellular automata to define a mathematical theory of computer simulations (Barrett & Reidys 1999), but this work has not yet considered inference from a collection of simulations. Our use of multiple simultaneous debates (scenarios) is *conceptually* similar to other work in AI using multiple possible worlds. For example, in the *Ents* model of belief of (Paris & Vencovská 1993), an agent's belief in a claim is determined by imagining possible worlds in which the claim is decided, either true or false, and then belief in the claim is set equal to the proportion of possible worlds in which it is true. In this model, the possible worlds are assumed equi-probable. This is also a feature of the model of (Bacchus et al. 1996), which assigns degrees of belief to propositions on the basis of the proportion of possible worlds in which there is evidence for them. In contrast, our approach allows scenarios (possible worlds) to be weighted differentially. Moreover, our approach provides a mechanism for deciding the truth-status of propositions within in each scenario, that of (McBurney & Parsons 2001b). Within AI, scenarios have also been used, e.g., as alternative possible explanations in probabilistic causal influence models (Henrion & Druzdzel 1991).

One criticism of the framework above is the possible sensitivity of conclusions to the particular weights assigned to scenarios. In future work we will seek to formalize the process of assigning ensemble weights, and to extend this overall approach beyond argumentation contexts. Potential applications will then include intelligent systems to aid decision-making in environmental domains (McBurney & Parsons 2001a), and assessment of scenario analysis in the climate change arena.

### Acknowledgements

We thank the anonymous referees for their comments